\documentclass{article} 
\usepackage{preprint,times}

\usepackage{amsmath,amsfonts,bm}

\def\eqref#1{equation~\ref{#1}}

\def\1{\bm{1}}

\DeclareMathAlphabet{\mathsfit}{\encodingdefault}{\sfdefault}{m}{sl}
\SetMathAlphabet{\mathsfit}{bold}{\encodingdefault}{\sfdefault}{bx}{n}

\usepackage{hyperref}
\usepackage{url}
\usepackage{amsfonts,amsmath}
\usepackage{nicefrac}
\usepackage{booktabs}
\usepackage{graphicx}
\usepackage{multirow}
\usepackage[font=small]{subcaption}

\hypersetup{
	colorlinks=true,%
	citecolor={blue!50!black},
	linkcolor={red!50!black},
	urlcolor={green!50!black}
}

\title{Learning Multi-Object Positional Relationships via Emergent Communication}

\author{Yicheng Feng \\
School of Computer Science\\
Peking University\\
\texttt{\small fyc813@pku.ecu.cn} \\
\And
Boshi An \\
School of Computer Science \\
Peking University \\
\texttt{\small boshi\_an@stu.pku.edu.cn} \\
\And
Zongqing Lu \\
Peking University\\
BAAI\\
\texttt{\small zongqing.lu@pku.edu.cn}
}

\iclrfinalcopy 

\begin{document}

\maketitle

\begin{abstract}
The study of emergent communication has been dedicated to interactive artificial intelligence. While existing work focuses on communication about single objects or complex image scenes, we argue that communicating relationships between multiple objects is important in more realistic tasks, but understudied. In this paper, we try to fill this gap and focus on emergent communication about positional relationships between two objects. We train agents in the referential game where observations contain two objects, and find that generalization is the major problem when the positional relationship is involved. The key factor affecting the generalization ability of the emergent language is the input variation between Speaker and Listener, which is realized by a random image generator in our work. Further, we find that the learned language can generalize well in a new multi-step MDP task where the positional relationship describes the goal, and performs better than raw-pixel images as well as pre-trained image features, verifying the strong generalization ability of discrete sequences. We also show that language transfer from the referential game performs better in the new task than learning language directly in this task, implying the potential benefits of pre-training in referential games. All in all, our experiments demonstrate the viability and merit of having agents learn to communicate positional relationships between multiple objects through emergent communication.
\end{abstract}

\section{Introduction}

In order to achieve interactive agents, a major problem to be solved is to endow artificial agents with the ability to communicate. Supervised methods are considered incapable of capturing functional meanings of language \citep{lazaridou2016multi, kottur2017natural}. Therefore, a series of studies on emergent communication probe into this problem by providing agents with simple environments where they learn to communicate with each other from scratch to accomplish specific tasks \citep{havrylov2017emergence, choi2018compositional, li2019ease, ren2020Compositional}. Most of these tasks are based on \textit{referential games} \citep{Lewis1969-LEWCAP-4}, where Speaker observes and describes a target object while Listener receives the message sent by Speaker and must pick out the target from several candidates.

In existing emergent language studies, agents' observations are mainly focused on a single object, be it a geometric object or a categorical image. Some studies involve images showing more complex scenes, but these studies usually also involve natural language \citep{das2017learning, DBLP:conf/nips/GuptaLL21}. Communicating the relationships between multiple objects explicitly is understudied. Then, problems may arise when we consider the development from communication in tasks like referential games to communication in tasks with more realistic settings, \textit{e.g.}, multi-step Markov decision process (MDP) tasks, since there the information about multi-object relationships is usually helpful, and sometimes even crucial. So in this paper, we try to fill this gap and address two questions: \textbf{\textit{Can neural agents learn to extract the information about multi-object relationships and express it through discrete communication channels in the referential game? If so, can the learned protocol help in more complex multi-step MDP tasks?}} We focus on positional relationships between two objects in this paper, because it is one of the most common and fundamental relationships, also usually most useful, and it is not too complicated, hence suitable as a starting point.

We train agents in the referential game where the observations are images each containing two geometric shapes, and see whether the agents can communicate the two objects and their positional relationship shown in each image. Since the positional relationship is abstraction information that can have various manifestations in specific images, we propose to use a \textit{random dataset} to test generalization, where each image is generated randomly each time, and the target image observed by Speaker and Listener is also different in pixel level but the same in abstraction. This is a stronger dataset than the standard setup, forcing agents to communicate abstract information to get high accuracy. We also use two common datasets as baselines, the \textit{fixed dataset} where images are fixed and the \textit{variation dataset} where images are randomly generated but the target image observed by Speaker and Listener is exactly the same. We find that agents trained with these two common datasets, though perform well if tested by the corresponding datasets, cannot generalize in the random dataset. This demonstrates that the two commonly used datasets cannot well test agents' ability to express abstract information, and also fail to help agents learn multi-object positional relationships. Instead, we find that agents trained with the random dataset can generalize well, implying that the input variation between Speaker and Listener is crucial for learning abstract information in emergent communication, so is necessary for extracting positional relationships. We also use an image encoder pre-trained by a contrastive learning method, SimCLR \citep{chen2020simple}, for comparison, and show that the language learned through the referential game with the random dataset generalizes better.

Then we show how communication about multi-object positional relationships helps in multi-step MDP tasks. We design a simple communication game where the positional relationship describes the goal. We find that the emergent language can generalize well in the new task, and is more powerful than raw-pixel images as well as pre-trained image features, proving the good generalization ability of discrete sequences. Besides, we find that language transfer from the referential game could achieve better performance than learning language from scratch in the new task, which may provide evidence for the benefits of language learning in the referential game.

We summarize the main contributions of our work as follows: (1) We explore agents' communication about multi-object positional relationships in raw-pixel images from scratch through emergent communication. (2) We propose to use the \textit{random dataset} to test the generalization of emergent languages, and find the environmental pressure where Listener observes target images different from Speaker's crucial for agents to emerge generalizable languages in the referential game. (3) Our experiments show that the emergent language can generalize well in the new multi-step MDP task, and is more powerful than raw-pixel images as well as pre-trained image features.

\section{Related Work}

\textbf{Emergent communication.} A series of studies have been done on emergent communication that trains interactive agents to learn protocols from communication games. Most studies focus on language learning in the referential game, where a speaker agent refers to targets using a message and a listener agent tries to understand the message \citep{lazaridou2016towards,lazaridou2016multi,lazaridou2018emergence,havrylov2017emergence,evtimova2018emergent,choi2018compositional,chaabouni2019anti,chaabouni2020compositionality,chaabouni2022emergent,dess2021interpretable,dagan2021co,DBLP:conf/nips/GuptaLL21}. These studies provide in-depth insights for learned protocols as well as learned representations of agents, but mostly stop at the single task. \citet{chaabouni2022emergent} proposed \textit{ease and transfer learning} (ETL) to evaluate the generalization of the emergent language to new tasks, but they do not involve multi-step MDP tasks.

Most studies exploring emergent communication in the context of the referential game use inputs containing a single object, \textit{e.g.}, a geometric shape or a natural image depicting a specific object. This restricts the generalization of the emergent language to complex MDP tasks. We go one step further to explore the positional relationship between two objects in observations.

Some other work explores emergent communication in multi-step MDP tasks directly, where agents learn to use discrete communication channels to cooperate \citep{bogin2018emergence,mordatch2018emergence,eccles2019biases,tucker2021emergent,lin2021learning}. These studies usually focus on methods for improving the ability of agents to accomplish the tasks through efficient communication, and explore whether the communication captures critical information for the tasks. However, the protocols are usually still specific to training tasks. We consider the generalization of the emergent language and probe into the language transfer from the referential game to more complex MDP tasks. And we think of the relationship between objects as an entry point.

\textbf{Input variation between Speaker and Listener in the referential game.} Most studies concerning the referential game use the same target input for Speaker and Listener. However, as \citet{bouchacourt2018how} mentioned, agents may fail to capture conceptual properties in inputs under this setup. \citet{mihai2019avoiding} augmented input images to Speaker with noise and random rotations to increase visual semantics of agents. \citet{lazaridou2016multi} and \citet{choi2018compositional} used a setup where Listener should choose a different image containing the same object as observed by Speaker to encourage the use of abstract information. Sharing the same idea, \citet{dess2021interpretable} used the data augmentation pipeline in SimCLR \citep{chen2020simple} to process input images. In our experiments, we find that adding noise alone is not enough for agents to communicate abstract information. We use a random image generator to introduce the environmental pressure more severely so that agents can almost never observe two same images. Moreover, we make a comparison with two other datasets, and find the random image generator really helpful for the communication about positional relationships.




\section{Experimental Setup}

\begin{figure}[t]
    \centering
    \setlength{\abovecaptionskip}{5pt}
    \includegraphics[width=.7\linewidth]{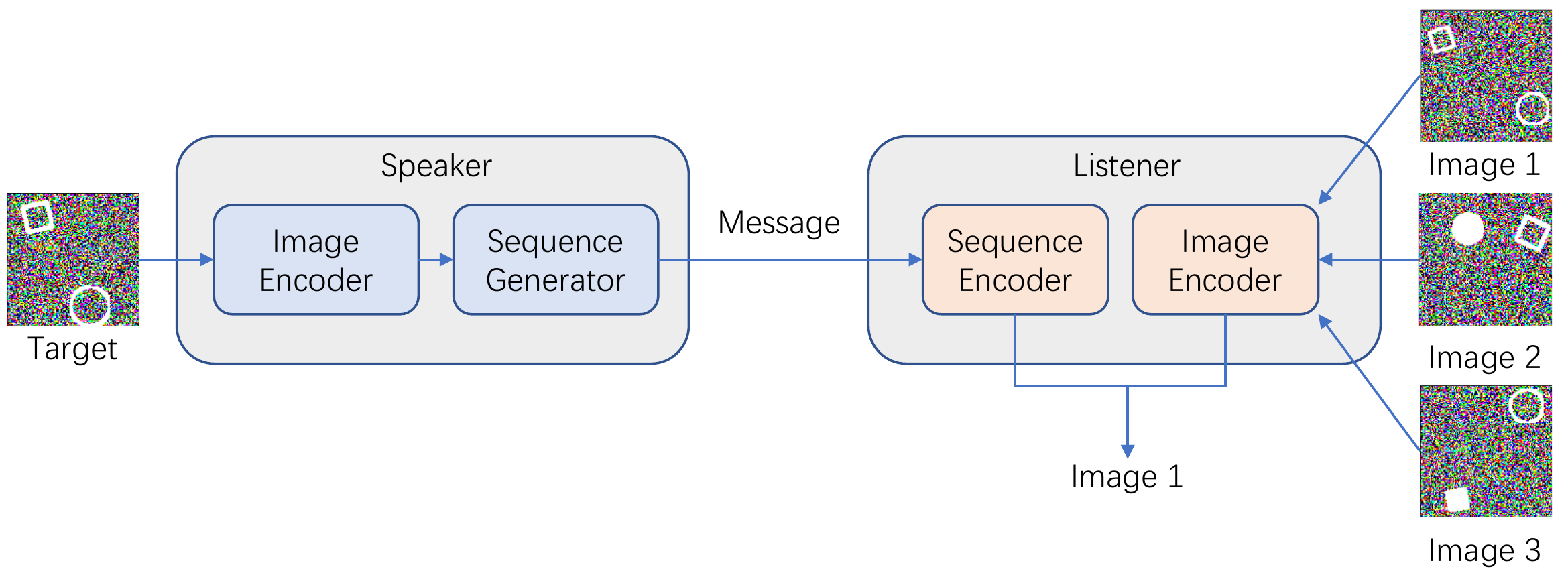}
    \caption{The referential game, agent architecture and examples of images in the random dataset.}
    \label{fig:referential game}
\end{figure}

\subsection{The referential game}
We train our agents in the two-player referential game where Speaker describes a target image to Listener who should pick out the target image among several candidates. Concretely, Speaker observes a target image $x$, and generates a message $m$ to describe it. The message $m$ is a sequence of discrete symbols from a vocabulary $\mathcal{V}$. The message length is $T$. Listener receives $m$ as well as a set of candidate images $\mathcal{C}$ including the target $x$ and several distractors. Then Listener selects an image $\hat{x}\in \mathcal{C}$ according to $m$. If $x = \hat{x}$, both agents get a reward $r=1$. Otherwise, the reward is 0. 

\subsection{Agent architecture}
Speaker, parameterized by $\theta$, consists of an image encoder and a sequence generator. The target image $x$ is first fed into a CNN network $f_\theta$ to get the image embedding $f_\theta(x)$. Then a projector $g_\theta$ maps the embedding into the initial hidden state of an LSTM \citep{hochreiter1997long}, $h_{-1}=g_\theta(f_\theta(x))$. Then at each time step $t$ a linear layer $\pi_\theta$ maps $h_t$ into a vector of dimension $|\mathcal{V}|$, and a symbol $w_t$ is sampled from the distribution induced by applying the softmax function to $\pi_\theta(h_t)$. And the one-hot embedding of the generated symbol $e(w_t)$ is fed back to the LSTM $l_\theta$ to update the hidden state $h_{t+1} = l_\theta(e(w_t), h_t)$. The first input symbol is a special token labeled as a start of sequence, $h_0 = l_\theta(e(sos), h_{-1})$. The symbols are generated until the message length reaches $T$. At test time, the symbols are not sampled but selected greedily.

Listener, parameterized by $\phi$, consists of an image encoder and a sequence encoder. An LSTM network $l_\phi$ encodes the sequence $m = w_0, w_1,...,w_{T-1}$ from Speaker into the message embedding $e_m = l_\phi(e(m))$, with each symbol in the sequence transformed to a one-hot embedding $e(m) = e(w_0), e(w_1),...,e(w_{T-1})$. A CNN network $f_\phi$ encodes each image $\Tilde{x} \in \mathcal{C}$ into image embedding $e_{\Tilde{x}} = f_\phi(\Tilde{x})$. A linear projector $p_{m,\phi}$ and an MLP projector $p_{\Tilde{x},\phi}$ projects the message embedding and each image embedding respectively to compute the cosine similarity between $p_{m,\phi}(e_m)$ and $p_{\Tilde{x},\phi}(e_{\Tilde{x}})$. The resulting similarities are passed to a softmax function to get a distribution over all images in the candidate set, and the image with the highest probability is selected. Details for hyper-parameters can be found in Appendix~\ref{appendix:architecture}.

\subsection{Datasets and the random image generator}
We create a dataset where we generate images of size $128\times 128$ each depicting two objects with a certain positional relationship between them. There are $5$ different objects and $4$ positional relationships (right, top right, top, and top left)\footnote{Due to the symmetry of the positional relationships, we do not include left, bottom left, bottom, and bottom right.}, so there are total \(100\) \((5\times5\times4)\) combinations. We use the word \textit{`combination'} to refer to the (object, object, relationship) tuple in the rest of the paper. We separate 20 of 100 combinations into the test set, so agents can only observe 80 combinations during training. We additionally add noise to the images for the robustness of the representation learning of image encoders, and to prevent degenerate policies of using pixel-level information. Accordingly, we set the message length $T=6$, the size of vocabulary $|\mathcal{V}|=5$, and the number of candidate images $|\mathcal{C}|=32$ for training and $|\mathcal{C}|=20$ for test in the referential game, which is illustrated in Figure~\ref{fig:referential game}.

In realistic environments, the observation of agents is ever-changing. So we propose to use a random image generator to generate specific images according to the combinations, where the absolute position, size, and orientation of objects vary. The details of the image generator can be found in Appendix~\ref{appendix:generator}. Then we hypothesize that using the random generator to provide images for Speaker and Listener separately can better test the generalization of agents, since agents can only succeed when they express and understand the abstract information in the images, especially when the multi-object positional relationship is involved because now images containing the same content are diverse at the pixel level.


To verify the hypothesis, we use other two kinds of datasets for comparison. Then we have three kinds of datasets as follows: (1) \textbf{Fixed dataset.} We do not use the random generator but generate one image for each combination, and the absolute position, size, and orientation of objects are fixed. This setup is similar to using structured input in some studies \citep{li2019ease,chaabouni2020compositionality,ren2020Compositional}, since there are no variations of each input in the dataset. Agents trained and tested with the fixed dataset can always observe only one instance of each combination. (2) \textbf{Variation dataset.} We use the random generator to generate images, but the target image observed by Speaker and Listener is the same one. This setup is similar to using natural images as inputs as in some studies \citep{chaabouni2022emergent,DBLP:conf/nips/GuptaLL21}, where different images depicting a same object exist in the dataset. Here agents see diverse images of a combination at training time, but may still use pixel-level information to succeed in the game. (3) \textbf{Random dataset.} We use the random generator and generate images for Speaker and Listener separately. Here agents almost never observe two same images and are forced to use abstract information to win the game.

\subsection{Optimization}

We use REINFORCE \citep{williams1992simple} to train Speaker which only uses the reward of the game. We also apply entropy regularization in the loss function to encourage exploration. To train Listener, we use the cross-entropy loss function which compares the output distribution of Listener with a one-hot vector indicating the target image. We use the default Adam optimizer \citep{kingma2015adam} with a learning rate of 3e-5 to update the parameters.

\section{Evaluation Methods}

\textbf{Generalization in referential games.} One of the most important properties of emergent language is the generalization ability to unseen inputs. We measure generalization in the referential game by the test accuracy.

\textbf{Compositionality.} We adopt a popular metric in emergent communication literature called \textit{topographic similarity (TopSim)} \citep{brighton2006understanding} for measuring language compositionality, which can also reflect generalization ability. It is computed by the Spearman correlation between the distances in the input space and the message space, so high \textit{TopSim} means that similar inputs lead to close messages. According to the characteristics of our setup, we compute the distance in the input space by the number of different attributes in the (object, object, relationship) tuple. We use the Levenshtein distance in the message space.

\textbf{Visual representations.} We explore the quality of the visual representations learned through the referential game. We focus on whether the representations contain features for abstract information, especially the positional relationship. Following \cite{dess2021interpretable}, we apply a linear projection head to the learned image encoder, and conduct a classification task trained by supervised learning on the test set. Then we use the classification accuracy to evaluate the learned visual representations.

\textbf{Ease and transfer learning (ETL).} \citet{chaabouni2022emergent} proposed \textit{ETL} to evaluate the generality of the emergent language to new Listener in new tasks. We measure \textit{ETL} by feeding the deterministic language (\textit{i.e.}, symbols are selected greedily) of Speaker to new Listener to perform new tasks and report the performances. We use two tasks for \textit{ETL}, image classification and Object Placement. The Object Placement task aims at our main research goal: whether and how the emergent language can generalize to multi-step MDP tasks.

\section{Experiments and Results}

\subsection{Input variation in the random dataset is important for communication about multi-object positional relationships}
\label{exp:referential}

\begin{figure}[t]
\setlength{\abovecaptionskip}{5pt}
  \centering
  \begin{subfigure}[t]{0.44\linewidth}
    \centering
    \includegraphics[width=.95\linewidth]{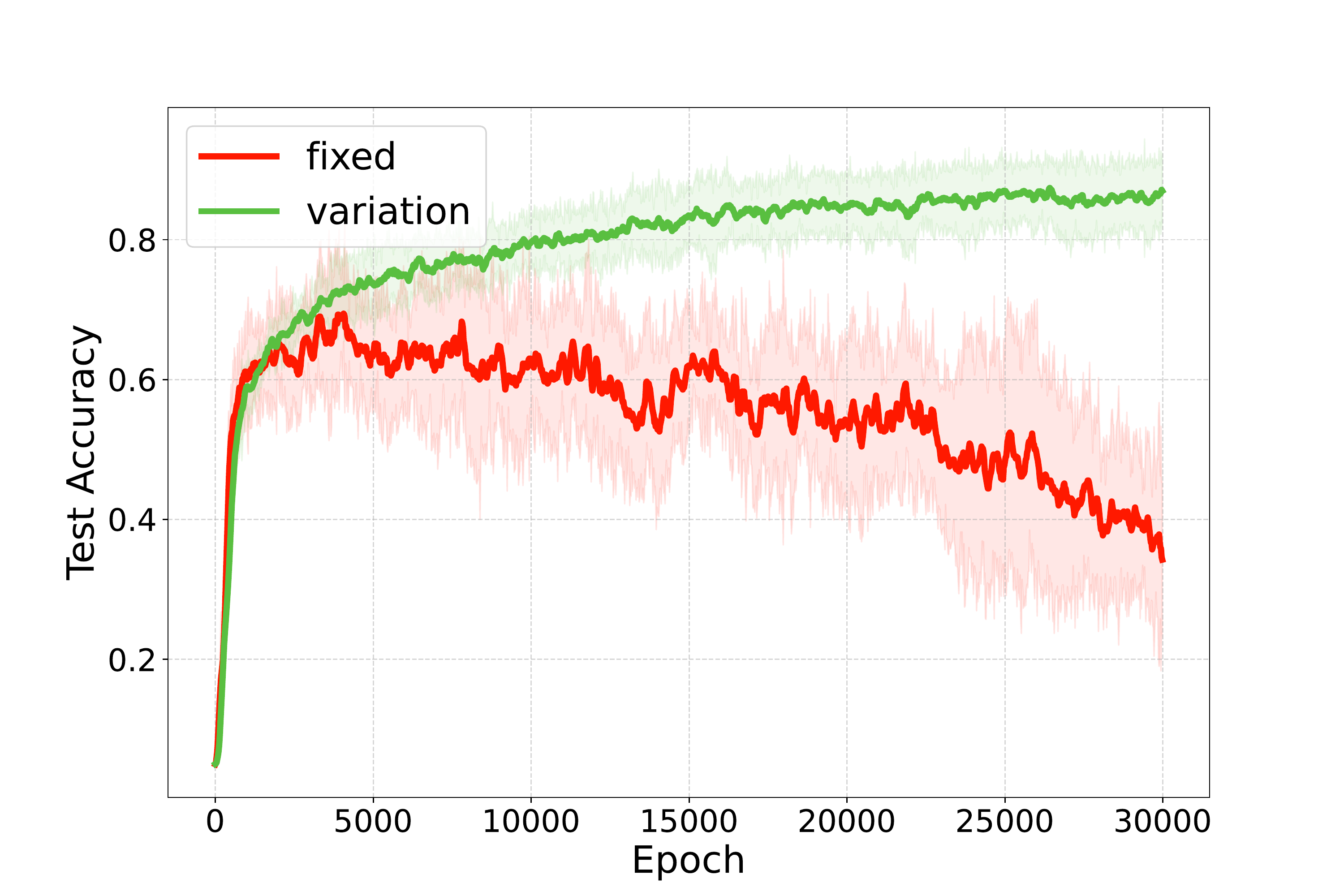}
    \caption{test accuracy on fixed and variation datasets}
    \label{fig:test_acc_other}
  \end{subfigure}
  \begin{subfigure}[t]{0.44\linewidth}
    \centering
    \includegraphics[width=.95\linewidth]{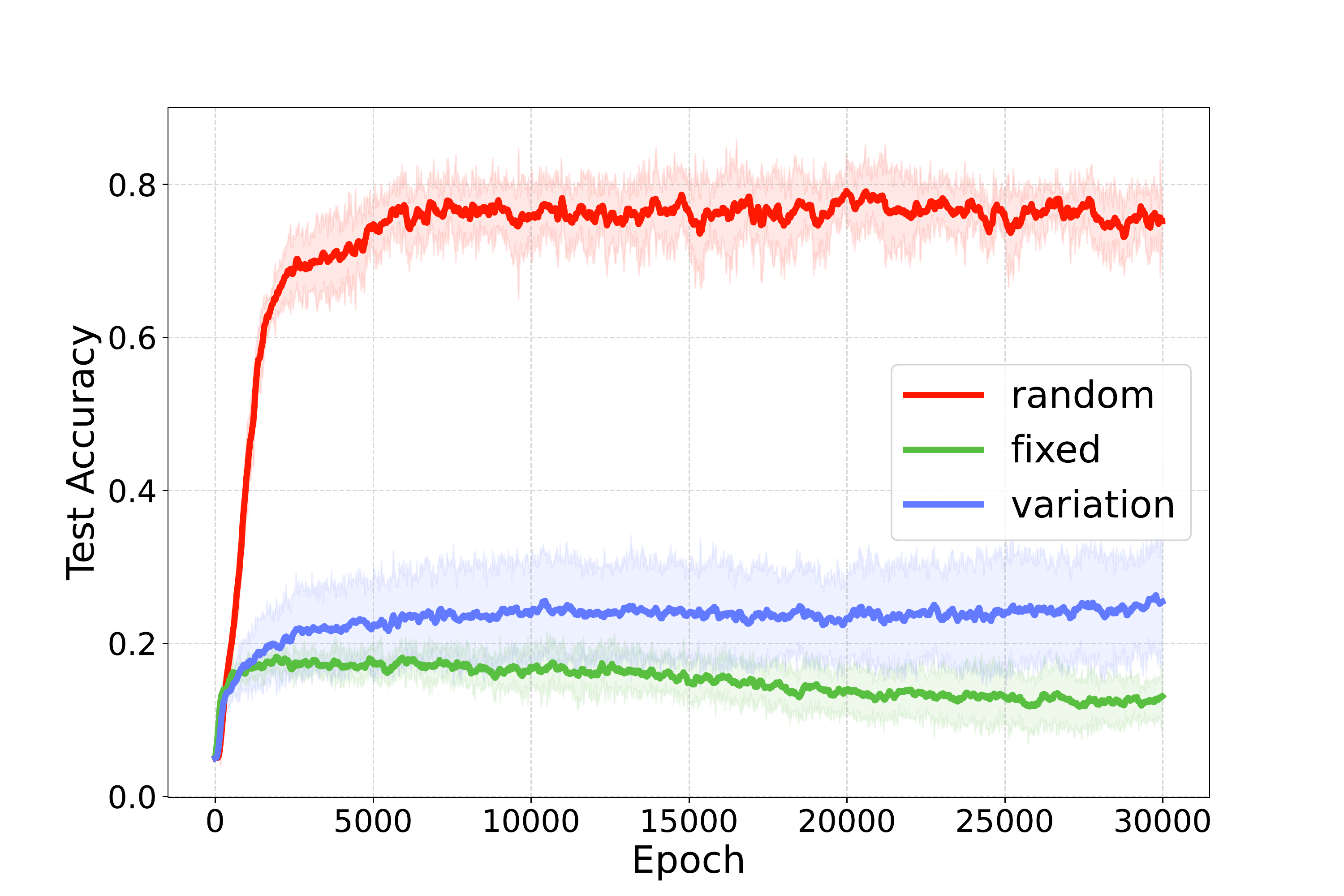}
    \caption{test accuracy on random dataset}
    \label{fig:test_acc_rand}
  \end{subfigure}
  \caption{Test accuracy of agents. (a) Agents trained with the fixed dataset or variation dataset are tested using the corresponding test set. (b) Agents trained with three kinds of datasets are tested with the test set of the random dataset.}
  \label{fig:test_acc}
\end{figure}

In this section, we analyze the performance of agents in the referential game learning to communicate the multi-object positional relationship from scratch. For all experiments, we run five times with different random seeds, and report the results in Figure~\ref{fig:test_acc}. We first use the fixed dataset and the variation dataset respectively for both training and testing. Results in Figure~\ref{fig:test_acc_other} show that agents trained with the variation dataset perform well at test time, so it seems to prove good generalization abilities. And agents trained with the fixed dataset can also get accuracies much higher than a random guess (5\%). However, when we use the random dataset for test, agents trained in the previous two datasets cannot generalize as shown in Figure~\ref{fig:test_acc_rand}. This implies that testing with the two commonly used datasets does not really reflect the generalization ability of agents. So we argue that input variation between Speaker and Listener is necessary for evaluating generalization in the referential game. Besides, agents trained in these datasets, though random noise is added, fail to communicate human-level conceptual information, at least when the positional relationship is involved.

\begin{figure}[h]
    \centering
    \setlength{\abovecaptionskip}{5pt}
    \includegraphics[width=.65\linewidth]{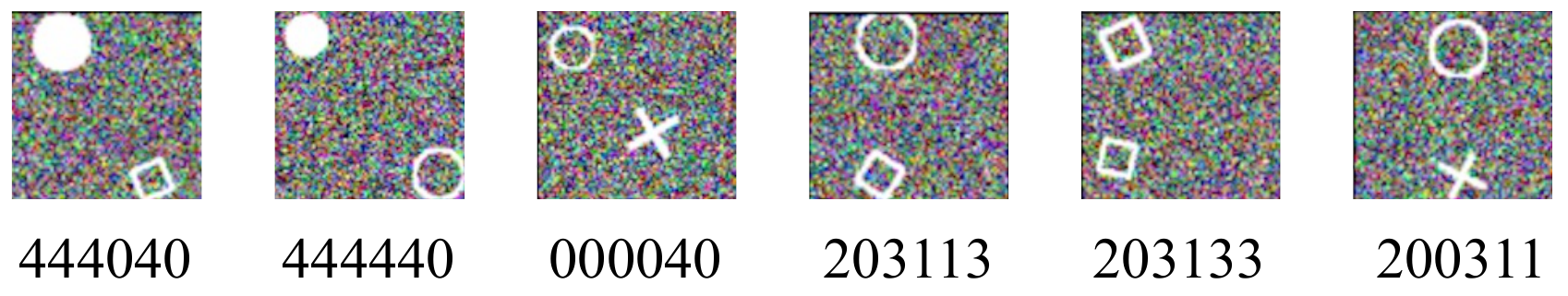}
    \caption{Examples of generated sequences by Speaker after training with the random dataset. The images are from the test set.}
    \label{fig:examples}
\end{figure}

Then how can agents learn to extract the positional relationship from images when communicating? A natural idea is to train agents with the random dataset, which provides a harsher environment. As mentioned in \citet{lazaridou2016multi} and \citet{choi2018compositional}, the input variation should encourage agents to use the abstract information. We show the results in Figure~\ref{fig:test_acc_rand}, and now the agents can perform well in the random dataset, with average accuracy close to 80\%. This proves that agents are communicating semantic information so Listener can understand and select the target even if the exact image is different from that observed by Speaker. So we argue that input variation between Speaker and Listener is also necessary for emergent communication about positional relationships, or even other abstract information, in the referential game. We present some examples of the generated sequences by Speaker observing images from the test set in Figure~\ref{fig:examples}. We can observe obvious patterns of different positional relationships in the sequences.

\begin{figure}[htb]
\vspace{-1mm}
\centering
\begin{minipage}[t]{0.44\linewidth}
  \setlength{\abovecaptionskip}{5pt}
    \centering
    \includegraphics[width=.95\linewidth]{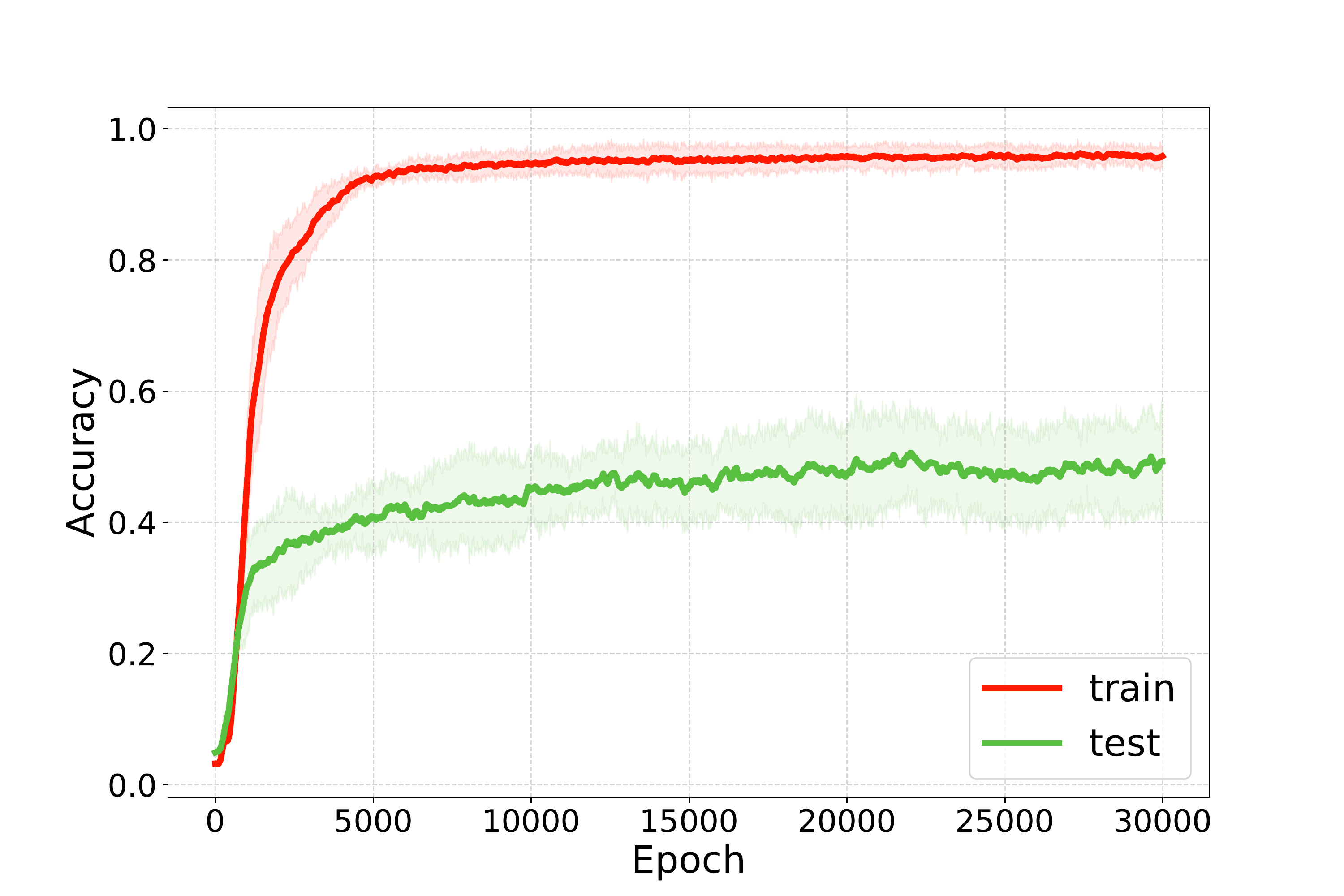}
    \caption{Train and test accuracy of agents whose image encoders are pre-trained by SimCLR with the random dataset.}
    \label{fig:acc_simclr}
\end{minipage}
\hspace{3mm}
\begin{minipage}[t]{0.44\linewidth}
    \centering
    \setlength{\abovecaptionskip}{5pt}
    \includegraphics[width=.95\linewidth]{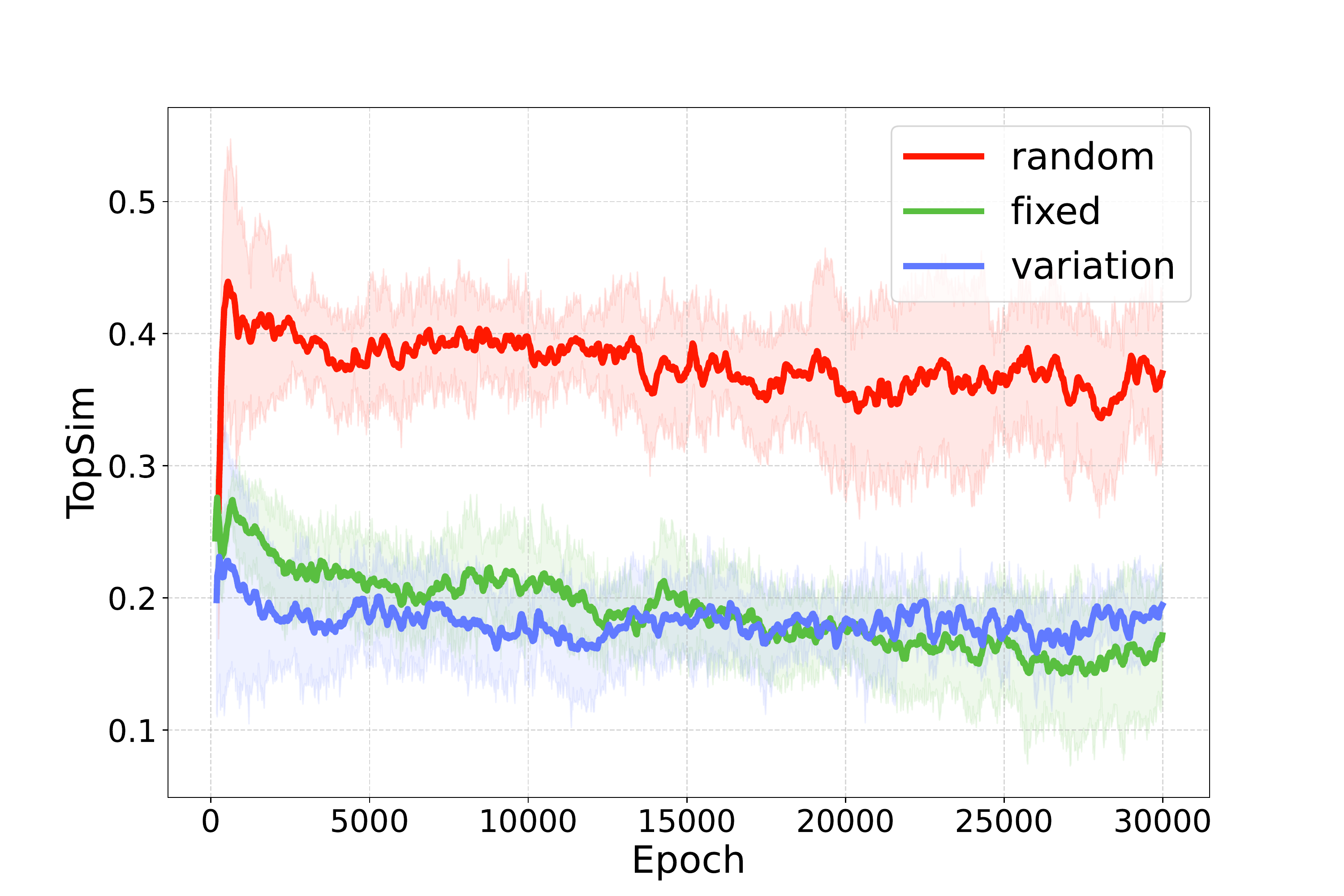}
    \caption{\textit{TopSim} of agents trained with different datasets.}
    \label{fig:topsim}
\end{minipage}
\end{figure}

\citet{dess2021interpretable} argues that the referential game is similar to the contrastive learning framework in SimCLR \citep{chen2020simple}. From this perspective, using the random dataset can be seen as a data augmentation process where the target image is changed but the semantic information is preserved. So we are curious about the performance of the representation learned with SimCLR instead of the referential game from scratch. We train a model using SimCLR, where the positive pairs are images generated by the random generator using the same combination in the training set. Then we use the frozen SimCLR model as pre-trained image encoders of Speaker and Listener, and train them in the referential game with the random training dataset. Finally, we test the agents using the random test dataset. The result is shown in Figure~\ref{fig:acc_simclr}. Surprisingly, using the pre-trained SimCLR model leads to worse performance compared to Figure~\ref{fig:test_acc_rand}, \textit{i.e.}, the agents cannot generalize well on the test set, though we find that they get a high accuracy at training time. 
One reason to explain the result may be that after SimCLR pre-training, the image representations of different images generated by the same combination are very similar, so the effect of using the random dataset in the following referential game is diminished, since the target representations observed by Speaker and Listener is almost the same now. From another perspective, the pre-trained encoders in advance separate different representations for different semantic information in the feature space, so the agents lose the environmental pressure to encode semantic information with emergent languages in the referential game, but can make use of some detailed information in the rich representation to accomplish the task. Then in the test set, though the pre-trained encoders can generate good representations for the new combinations, the agent language cannot generalize well to the new representations.
This result shows that using pre-trained image encoders may do bad to generalization in emergent communication.

\subsection{Analysis of protocols and representations learned through the referential game}
\label{sec:visual rep}

We report the results for computing \textit{TopSim} for agents trained with different datasets in Figure~\ref{fig:topsim}. Obviously, agents trained with the random dataset get higher \textit{TopSim}, so they tend to use similar messages to describe similar inputs, implying more compositional languages. This again demonstrates the benefit of using the random dataset for training.

\begin{table}[tbh]
\caption{We report the mean classification accuracy on our test set with images generated by the random image generator of five different seeds, and one standard error in the brackets. The first row is the evaluation of Speaker's visual representations trained with different datasets as discussed in Section~\ref{sec:visual rep}. The second row is the image classification task of \textit{ETL} as illustrated in Section~\ref{etl_1}.}
\label{visual-rep-etl1}
\vspace{-2mm}
\begin{center}
\begin{tabular}{llll}
\toprule
& Fixed & Variation & Random\\
\midrule
Visual representation \scriptsize{(\%)} & 84.4 \scriptsize{(5.5)} & 76.2 \scriptsize{(8.9)} & \textbf{100.0} \scriptsize{(0.0)}\\
ETL-image classification \scriptsize{(\%)} & 32.8 \scriptsize{(8.1)} & 31.8 \scriptsize{(7.2)} & \textbf{90.8} \scriptsize{(2.8)}\\
\bottomrule
\end{tabular}
\end{center}
\end{table}

Then we evaluate Speaker's visual representations learned through the referential game. We conduct a classification task to examine whether the visual representations encode conceptual information. We apply a linear classifier to the frozen CNN of Speaker and train it on our test set with images generated by the random image generator. 
Results in Table~\ref{visual-rep-etl1} demonstrate that agents trained with the random dataset learn better visual representations that capture conceptual information, and perform perfectly in the classification task on the test set. This shows us a promising direction that the referential game can serve as a good representation learning approach that may help encode high-level abstract information in features. On the other hand, the variation dataset does not perform better than the fixed dataset, so the key factor influencing the quality of visual representations is the input variation between Speaker and Listener instead of variations in the dataset. Since representation learning plays an important role in emergent communication, the result tells us that input variation between Speaker and Listener should get attention.

\subsection{Language generalization in new tasks}

We adopt \textit{ETL} proposed in \citet{chaabouni2022emergent}, which is considered a more robust metric, to evaluate the ability of the emergent language to generalize to new Listener and new tasks. We conduct a image classification task in Section~\ref{etl_1} as in \citet{chaabouni2022emergent}. Moreover, we want to extend the new tasks to more complex multi-step MDP tasks, which can hardly be achieved if agents can only refer to single objects. We explore this with a task named Object Placement in Section~\ref{etl_2}.

\subsubsection{Image classification}
\label{etl_1}
For the image classification task, We feed the deterministic language of Speaker to new Listener and train a linear classifier on the hidden state of Listener's sequence encoder on our test set with images generated by the random image generator. The results are shown in Table~\ref{visual-rep-etl1}. We can find that \textit{ETL} faithfully reflects the generalization ability of agents, with the random dataset showing the best performance. On the other hand, since \textit{ETL} focuses on the information content conveyed by Speaker, the result implies that agents trained with the random dataset can express the positional relationship well. Note that the combinations are never seen by Speaker in the referential game, and the random image generator provides totally different images of the same content, but new Listener can easily understand the messages and achieve the classification accuracy over 90\%, proving that Speaker has already learned to convey the conceptual information in images. Contrarily, agents trained with the fixed dataset and variation dataset cannot learn to communicate such information clearly. So in general, we can conclude that agents can learn to communicate multi-object positional relationships through emergent communication, but necessary environmental pressure should be involved, such as the input variation between Speaker and Listener.

\begin{figure}[t]
  \setlength{\abovecaptionskip}{5pt}
    \centering
    \includegraphics[width=.9\linewidth]{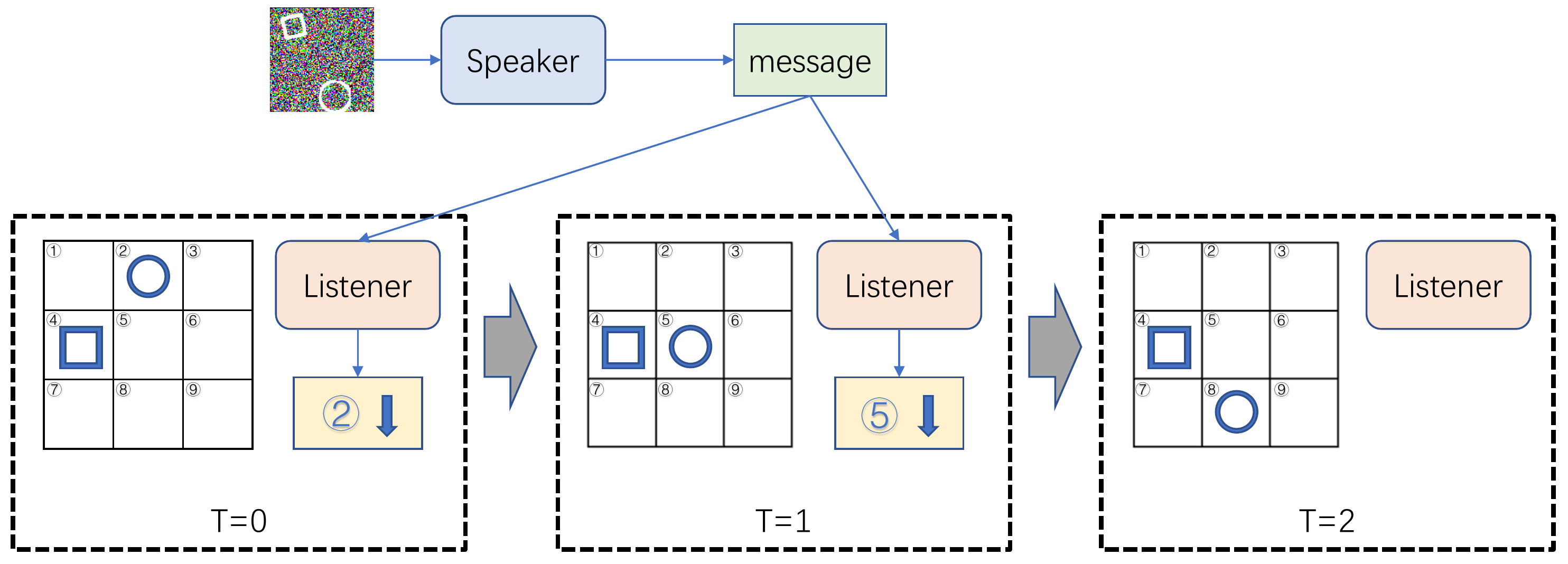}
    \caption{Object Placement task. Speaker observes the target state (image) and describes it to Listener. Listener observes the grid world containing the two objects and receives the message from Speaker. Then it moves the objects to place them to form the correct positional relationship as depicted in the target state.}
    \label{fig:mdp_task}
\end{figure}

\subsubsection{Object Placement}
\label{etl_2}
Now, according to the analysis above, we have addressed the first question that agents can learn to express positional relationships in the context of the referential game. Then we explore the second one: whether the learned protocol can be helpful in multi-step MDP tasks with the ability to convey information about positional relationships. We design a task named Object Placement, as illustrated in Figure~\ref{fig:mdp_task}. Speaker observes a target image depicting the target positional relationship of two objects. It then sends a message to Listener, who should move the objects in the $3\times 3$ grid to place them in the corresponding positional relationship. The action of Listener is to choose a grid and a direction, and if there is an object in the grid, the object is moved according to the direction by one grid. The observation of Listener is the state of the grid world and the message sent by Speaker. 
If Listener places two objects in the correct positional relationship, the reward is $+1$ and the episode terminates, otherwise, the reward is $-0.01$ for each step. The maximum episode length is set to $20$.
The target images are sampled from our training set generated by the random image generator. We use Speaker trained with the random dataset in the referential game, and generate deterministic messages to Listener. Listener uses a newly initialized sequence encoder to process the messages. We train Listener with PPO \citep{schulman2017proximal}.

We also compare with five baselines: 
\begin{itemize}
    \item The \textit{raw-pixel-input} baseline uses target images to replace the messages sent by Speaker, and Listener learns a CNN model to process the images; 
    \item The \textit{cnn-feature} baseline also uses target images to replace the messages, but Listener uses a frozen CNN model pre-trained on our training set with the random generator by an image classification task;
    \item The \textit{simclr-feature} baseline uses a pre-trained SimCLR model instead of the pre-trained CNN model compared with the \textit{cnn-feature} baseline;
    \item The \textit{rl-scratch} baseline trains Speaker from scratch using REINFORCE to send messages. For this method, we train Speaker and Listener alternately;
    \item The \textit{state} baseline gives the true target relation to Listener directly, showing the optimal performance. 
\end{itemize}
Details for the Object Placement task and the baselines can be found in Appendix~\ref{appendix: object placement}.

\begin{figure}[t]
\setlength{\abovecaptionskip}{5pt}
  \centering
  \begin{subfigure}[t]{0.44\linewidth}
    \centering
    \includegraphics[width=.95\linewidth]{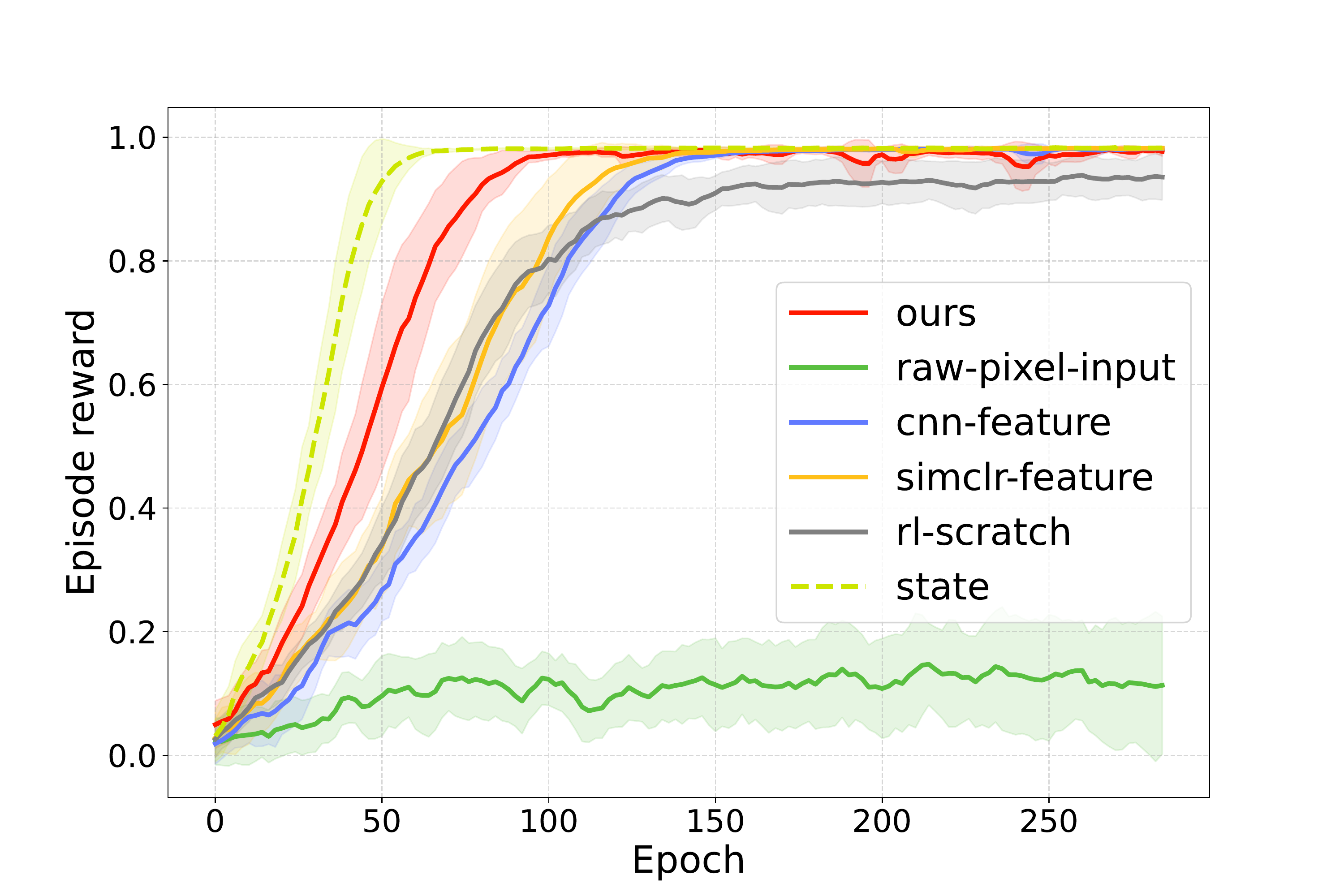}
    \caption{episodic reward in Object Placement task}
    \label{fig:ep_rew}
  \end{subfigure}
  \hspace{0.2cm}
  \begin{subfigure}[t]{0.44\linewidth}
    \centering
    \includegraphics[width=.95\linewidth]{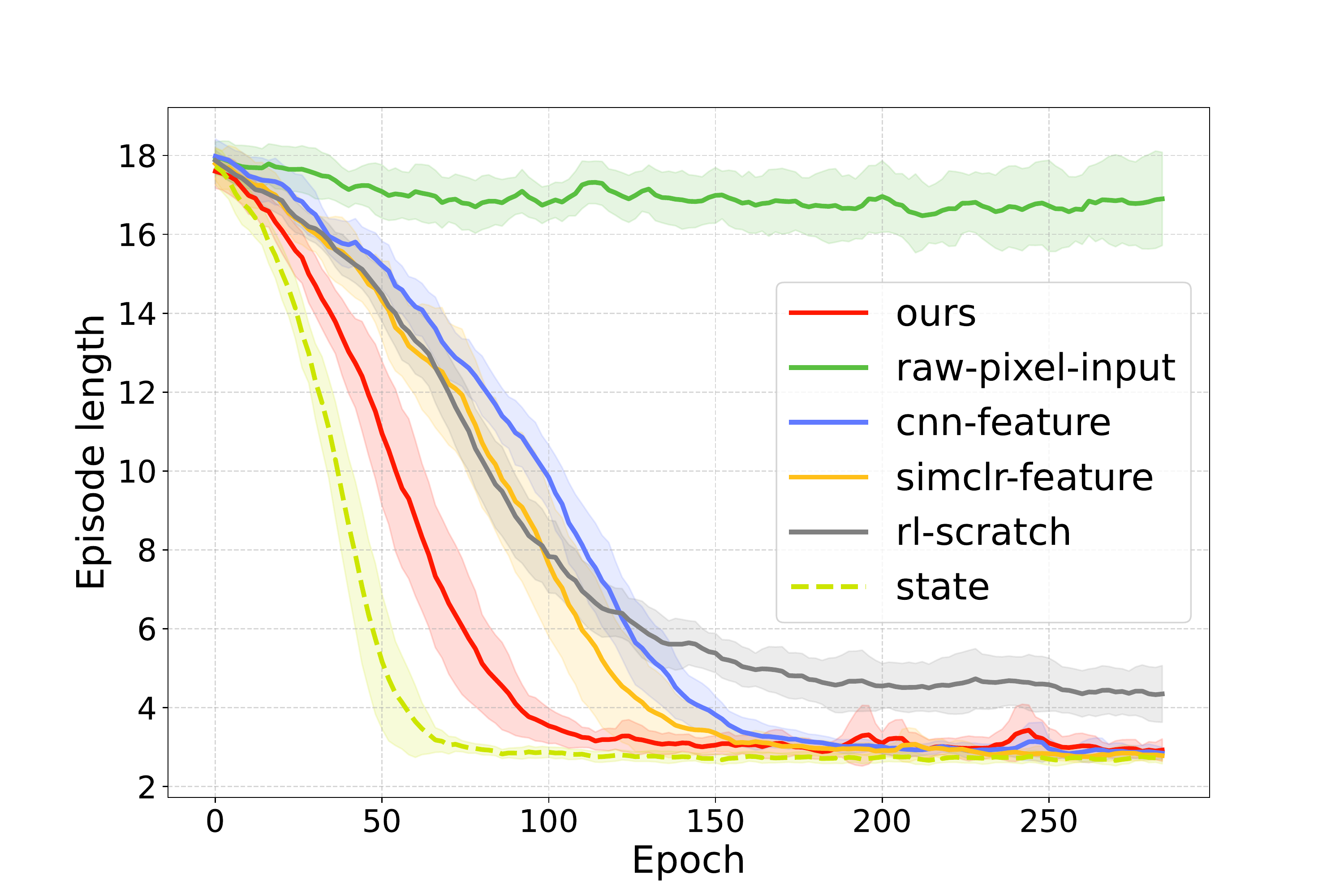}
    \caption{episode length of accomplishing the task}
    \label{fig:ep_len}
  \end{subfigure}
  \caption{Performance of new Listener trained with different inputs of the target state in the Object Placement task. All experiments are run for 5 seeds, and the shaded part of the curves is one standard error.}
  \label{fig:obj}
\end{figure}

Figure~\ref{fig:obj} shows the learning curves of all the methods in the Object Placement task: the episode reward in Figure~\ref{fig:ep_rew}, and the episode length of agents accomplishing the task in Figure~\ref{fig:ep_len}. Except the \textit{rl-scratch} and \textit{raw-pixel-input} baselines, all other methods converge to the same performance but differ in learning speed. 

Firstly, from the \textit{ETL}'s perspective, our Speaker's language can generalize pretty well in the new multi-step task, so new Listener can understand the message and learn a good policy in the new task quickly, close to the \textit{state} baseline (the upper bound) that tells Listener the true target relationship. This demonstrates the generalization ability of the emergent language in the referential game, and shows that the agent has learned a general communication skill instead of a protocol overfitting to a single task. And this addresses our second question that emergent language in the referential game can be helpful in multi-step MDP tasks. Previous studies where agents learn to refer to single objects hardly explore the language transfer to multi-step tasks, probably because the object-level information is usually not sufficient for accomplishing these tasks. Our research on the learning of positional relationships can be seen as a step to break the restriction and towards the application of emergent communication in more complex tasks.

Besides, the \textit{raw-pixel-input} baseline fails to learn a policy to accomplish the task. This result proves that agents trained with deep reinforcement learning may feel difficult to capture the abstract information from raw-pixel images directly, so the Listener seems confused with this input. Therefore, state representations become important for reinforcement learning agents when the environment requires abilities for conceptual abstraction.

Then which kind of representation is better? In Figure~\ref{fig:obj} we can find that, though the \textit{cnn-feature} baseline and the \textit{simclr-feature} baseline achieve comparable performance with our method that uses the learned Speaker, Listener learns faster if the input is discrete symbols. This is to some extent in line with the point of view in \citet{garnelo2016towards} that conceptual abstraction provided by symbolic representations promotes data efficient learning. So it comes to the significance of research on language learning about conceptual information that is useful in various MDP tasks, such as positional relationships, spatial relationships, or numeric concepts \citep{guo2019emergence}.

From the result of the \textit{rl-scratch} baseline, we find that training Speaker and Listener directly in the Object Placement task gets poorer performance than using pre-trained emergent language. This may provide evidence that the referential game is more suitable to serve as a starting point for language learning, since it is easier for compositional and generalizable languages to emerge. It is reasonable because in the referential game Speaker receives the feedback more effectively.


\begin{figure}[t]
\setlength{\abovecaptionskip}{5pt}
  \centering
  \begin{subfigure}[t]{0.55\linewidth}
    \centering
    \includegraphics[width=.76\linewidth]{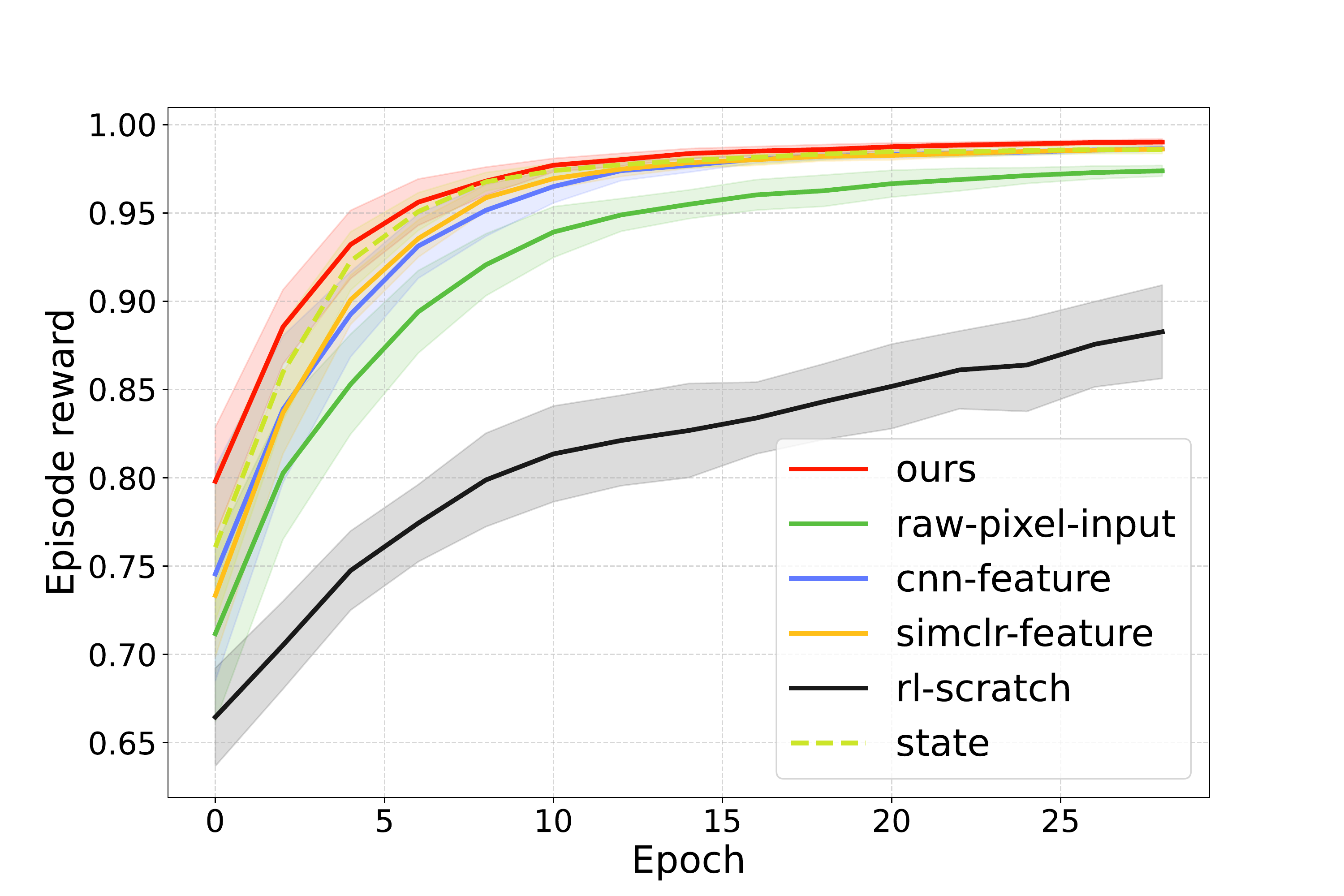}
    \caption{episodic reward in Multi-Listener Object Placement task}
    \label{fig:multi_ep_rew}
  \end{subfigure}
  \hspace{0.2cm}
  \begin{subfigure}[t]{0.418\linewidth}
    \centering
    \includegraphics[width=1\linewidth]{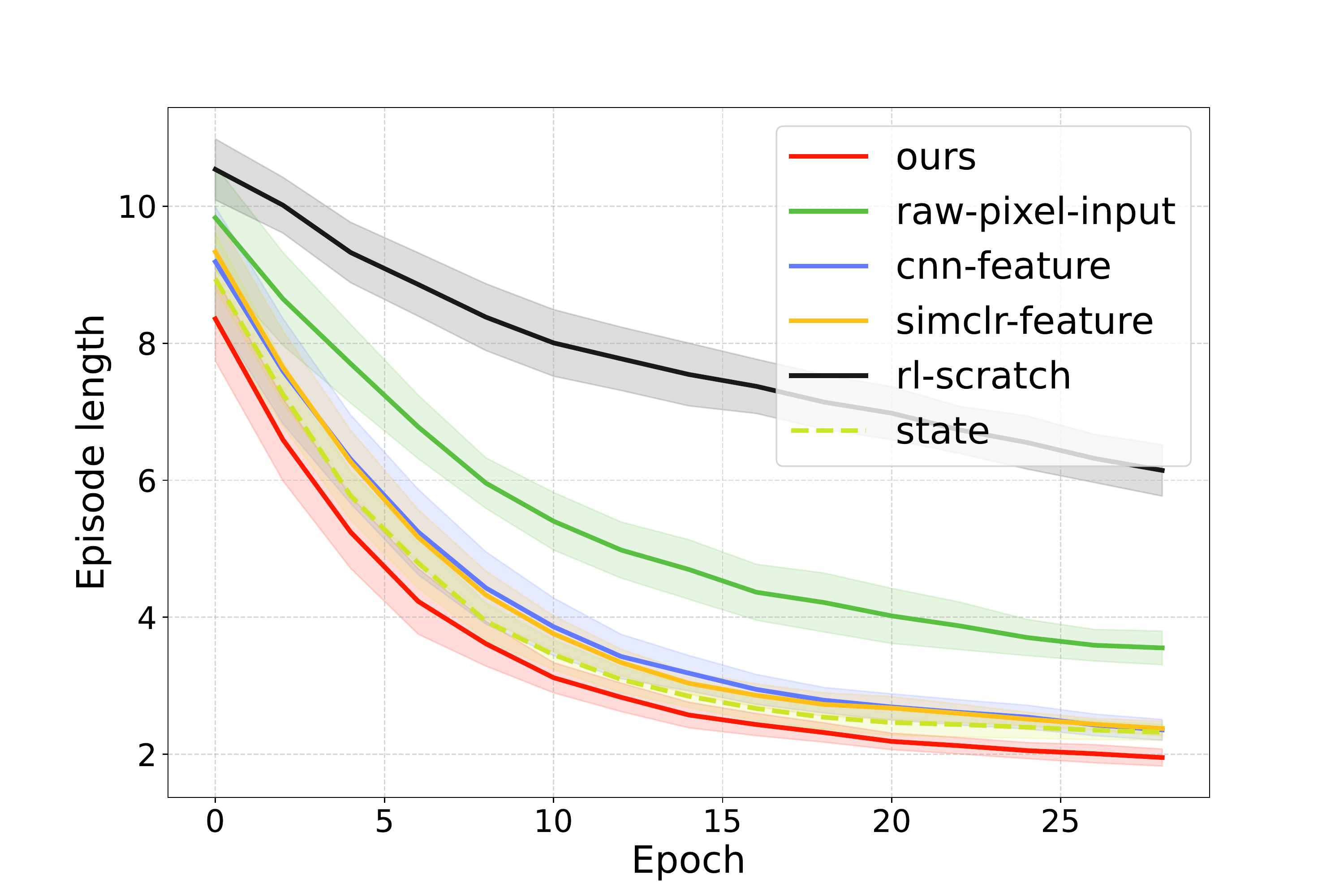}
    \caption{episode length of accomplishing the task}
    \label{fig:multi_ep_len}
  \end{subfigure}
  \caption{Performance of new Listeners trained with different inputs of the target state in the Multi-Listener Object Placement task. All experiments are run for 5 seeds, and the shaded part of the curves is one standard error.}
  \label{fig:obj_multi}
\end{figure}

We then extend the task to scenarios involving multiple Listeners to test the robustness of the generalization ability of the emergent language. We modify the Object Placement task to the Multi-Listener Object Placement task where we now have two independent Listeners each can move one object in the grid world. Then they should cooperate to achieve the goal. Both Listeners receive the same observation containing the state of the grid world and the message from Speaker, and the action is the moving direction of the object they control. We also compare with the baselines in the last experiment. The results are shown in Figure~\ref{fig:obj_multi}.

While all methods except the \textit{rl-scratch} can perform well, our method using the learned symbolic language still learns faster, even compared with the goal state input. So the generalization ability of the emergent language is also effective when multiple new Listeners learn to understand the language at the same time. This shows the robustness of our finding that emergent language can be used in various MDP tasks thanks to its good generalization ability, and increasing its expressive power expands the range of its application in MDP tasks.

\section{Discussion}
The goal of emergent communication should be making neural agents acquire general communication skills instead of merely the ability to solve specific communication games. Many studies have been dedicated to the research on learning compositional languages in the context of referential games, but few have probed into the generalization of the emergent language to more complex tasks such as multi-step MDP tasks. We wonder about the viability of this development, while we argue referential games restricted to referring to single objects limit such development. So we go one step forward to explore communication about positional relationships, which may be an entry point of emergent communication about more high-level conceptual information.

We first find that agents can learn to communicate positional relationships well through training with the referential game, but the key factor that influences the ability is the input variation between Speaker and Listener. So we may need stronger environmental pressure when more conceptual information is involved. We also show that we need stronger datasets to test the true generalization ability of emergent languages. 

Then we use a simple environment to evaluate the performance of language transfer from the referential game to a multi-step MDP task. We find that the emergent language, which can convey information about positional relationships, not only generalizes well in the new task, but also overperforms pre-trained image features and language learned directly in the specific task. So it verifies the viability of language transfer from referential games to more complex tasks, and shows a promising path to employ emergent communication for conceptual abstraction in complex environments and games.

It is worth noting that we focus on learning positional relationships in the referential game in this paper, and we have carried out preliminary experiments of language transfer from the referential games to complex MDP tasks. The limitations in this work should be addressed in future: whether, or how, the learned positional relationships can generalize well to out-of-distribution datasets? Then the acquired communication skills can be applied to more diverse tasks. Besides, the Object Placement task in our work is somewhat simple, and we should explore language transfer to more general MDP tasks in future work. Furthermore, positional relationship is not enough for general tasks, whether other conceptual information can be learned through emergent communication? In addition to serving as a function similar to state representation, grounding the emergent language into actions in MDP tasks is also a future direction. Our work may be seen as one of the openings for research on task scaling up for more general agent language learning through emergent communication.




\bibliography{preprint}
\bibliographystyle{preprint}

\newpage
\appendix
\section{Agent Architecture and Hyperparameters}
\label{appendix:architecture}
\textbf{Speaker architecture}

Speaker consists of an image encoder and a sequence generator.

\begin{enumerate}
    \item The image encoder $f_\theta$ is a reduced AlexNet, receiving images of size $128 \times 128$ and outputs embeddings of size $216$.
    \item A projector $g_\theta$ maps the embedding $f_\theta(x)$ into the initial hidden state of the sequence generator, composed of a Linear layer with input size of 216 and output size of 128 and a ReLU activation.
    \item The sequence generator is an LSTM network with hidden size 128.
    \item A Linear layer $\pi_\theta$ with input size of 128 and output size of $|\mathcal{V}|$ maps the hidden state of the sequence generator $h_t$ into a logits vector, which is then fed to a softmax function to produce the symbol distribution.
\end{enumerate}

\textbf{Listener architecture}

Listener consists of an image encoder and a sequence encoder.

\begin{enumerate}
    \item The architecture of the image encoder $f_\phi$ is the same as that of Speaker $f_\theta$, and the parameters are not shared across the Listener and the Speaker.
    \item The sequence encoder $l_\phi$ is an LSTM network with hidden size 256. It receives one-hot embeddings of symbols.
    \item The MLP projector $p_{\Tilde{x},\phi}$ is composed of a Linear layer with input size 216 and output size 128, a ReLU activation, and a Linear layer with input size 128 and output size 128.
    \item The linear projector $p_{m,\phi}$ is a Linear layer with input size 256 and output size 128.
\end{enumerate}

\textbf{Other hyper-parameters}

The batch size and the candidate number $|\mathcal{C}|$ in the referential game are set to 32 for training and 20 for testing. The learning rate is 3e-5, and the entropy coefficient is 0.01.

The batch size for classification tasks in Section~\ref{sec:visual rep} and Section~\ref{etl_1} is 128. We use default Adam optimizer here with learning rate 3e-4.

Listener in the Object Placement task is trained using default setups of PPO algorithm of the stable-baselines3 repository \citep{stable-baselines3}.
\section{Random Image Generator}
\label{appendix:generator}

The random image generator takes three input parameters: two of them describing the shape of the objects and one for the positional relation between the objects, and generates an image according to the parameters. We added randomization to the size, rotation and position of each object. More precisely, the size (pixels) of each object is a random variable sampled from the interval $[28, 40]$ uniformly and independently, the rotation angle is uniformly sampled from $0$ to $359$ degrees. If the required positional relation is right, the horizontal displacement from one object to the other is uniformly sampled from the interval $[50, 88]$ and the vertical displacement is uniformly sampled from the interval $[-5, 5]$ . If the required positional relation is top right, both the horizontal and vertical displacement is sampled from the interval $[50, 88]$, uniformly and independently. If the required positional relation is top, the horizontal displacement is uniformly sampled from the interval $[-5, 5]$ and the vertical displacement is uniformly sampled from the interval $[50, 88]$ . If the required positional relation is top left, the horizontal displacement is uniformly sampled from the interval $[-88, -50]$ and the vertical displacement is uniformly sampled from the interval $[50, 88]$ . The background of the image generated is black ($R=G=B=0$) and the shapes are colored white ($R=G=B=255$). A noise sampled from $\mathcal{N}(0, 16)$ is added to each channel of pixels.


\section{Object Placement}
\label{appendix: object placement}

\subsection{The Object Placement MDP task }\label{mdp_description}
The object placement MDP task consists of a $3\times 3$ grid with two objects in two different grids. The speaker is given the image of the target state of the MDP environment, while the listener is given the state-based description of each object, and need to move the objects to reach the target state. More precisely, the image given to speaker is generated by the random image generator described in Appendix~\ref{appendix:generator}, the observation for the listener contains a 6-element tuple consisting the X,Y coordinate and shape index of each object and the output sequence from the speaker. At each time step, the speaker describes the image, and the output sequence is given to the listener along with the state-based observation of the MDP environment. The listener then selects a grid (represented by its coordinate) and a direction (right, left, up and down) as the action, which means the object on the selected grid should be moved to the adjacent grid in the selected direction. The move will be successfully applied to the environment if there is an object in the selected grid and the target grid of current movement is empty. The reward of each move is either $1.0$ or $-0.01$ . The reward is $1.0$ when the positional relationship between the two objects in the MDP environment is the same as which in the image given to the speaker, otherwise, the reward is $-0.01$ as a penalty.

\subsection{The Multi-Listener Object Placement task}
We modify the Object Placement task to involve multiple Listeners. Concretely, all the settings are the same as in \ref{mdp_description}, except that there are two listeners and the action space is different from the original task. At each step, both listeners are given the same observation, which contains a 6-element tuple consisting the X,Y coordinate and shape index of each object and the output sequence from the speaker. Each listener then selects a direction (right, left, up and down) as the action. The action of the first listener will only control the moving direction of the first object while the action of the second listener will only control the moving direction of the second object.

\subsection{Baselines}

\textbf{Raw-pixel-input}

This baseline provides the target image directly to Listener, and Listener use a CNN network to process the input image.

\textbf{CNN-feature}

The CNN network architecture for this baseline is the same as the image encoder of Listener in the referential game. We pre-train it on our training set using the random image generator with a classification task. We consider each of the 80 combinations as a class, and use cross-entropy loss to train the network. We apply an MLP projector to the output feature when pre-training, and the projector is abandoned in the Object Placement task.

\textbf{SimCLR-feature}

This baseline is similar to the CNN-feature baseline, but the CNN network is pre-trained with SimCLR method, as described in the third paragraph in Section~\ref{exp:referential}.

\textbf{RL-scratch}

We train Speaker and Listener alternatively. When training Speaker, Listener is part of the environment. In each episode, Speaker produces a message, and we argmax Listener's policy to get actions in Object Placement task for the episode, computing the reward. We then train Speaker using REINFORCE with entropy regularization. When training Listener, we fix Speaker and the process is the same as our method. In each phase we train 1000 steps for both Speaker and Listener.

\textbf{State}

This baseline directly provides the current state and target state to the Listener, and Listener use a MLP network to process the input states.

\section{Additional Results}
\begin{figure}[t]
\setlength{\abovecaptionskip}{5pt}
  \centering
  \begin{subfigure}[t]{0.44\linewidth}
    \centering
    \includegraphics[width=.95\linewidth]{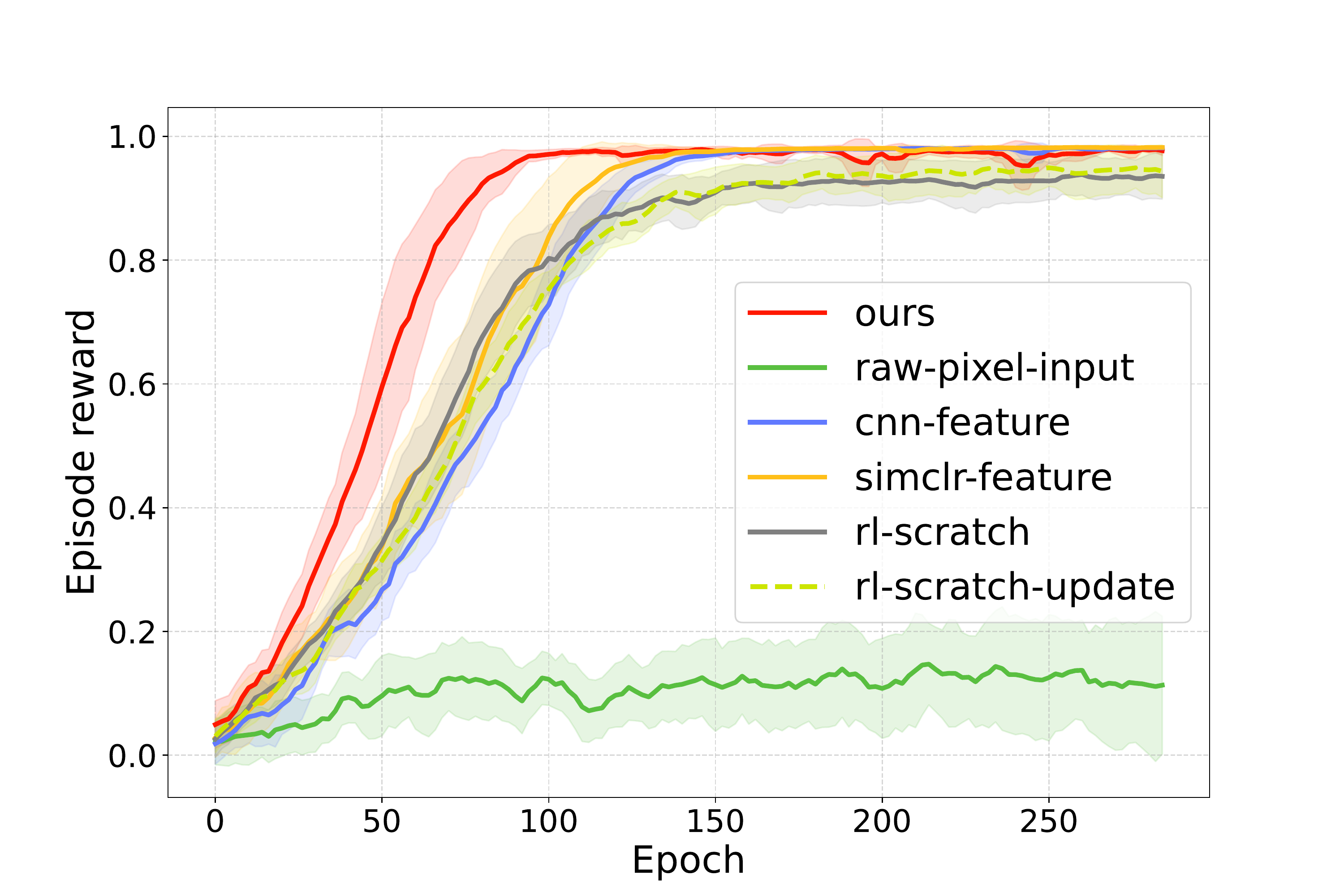}
    \caption{episodic reward in Object Placement task}
    \label{fig:app-ep_rew}
  \end{subfigure}
  \hspace{0.2cm}
  \begin{subfigure}[t]{0.44\linewidth}
    \centering
    \includegraphics[width=.95\linewidth]{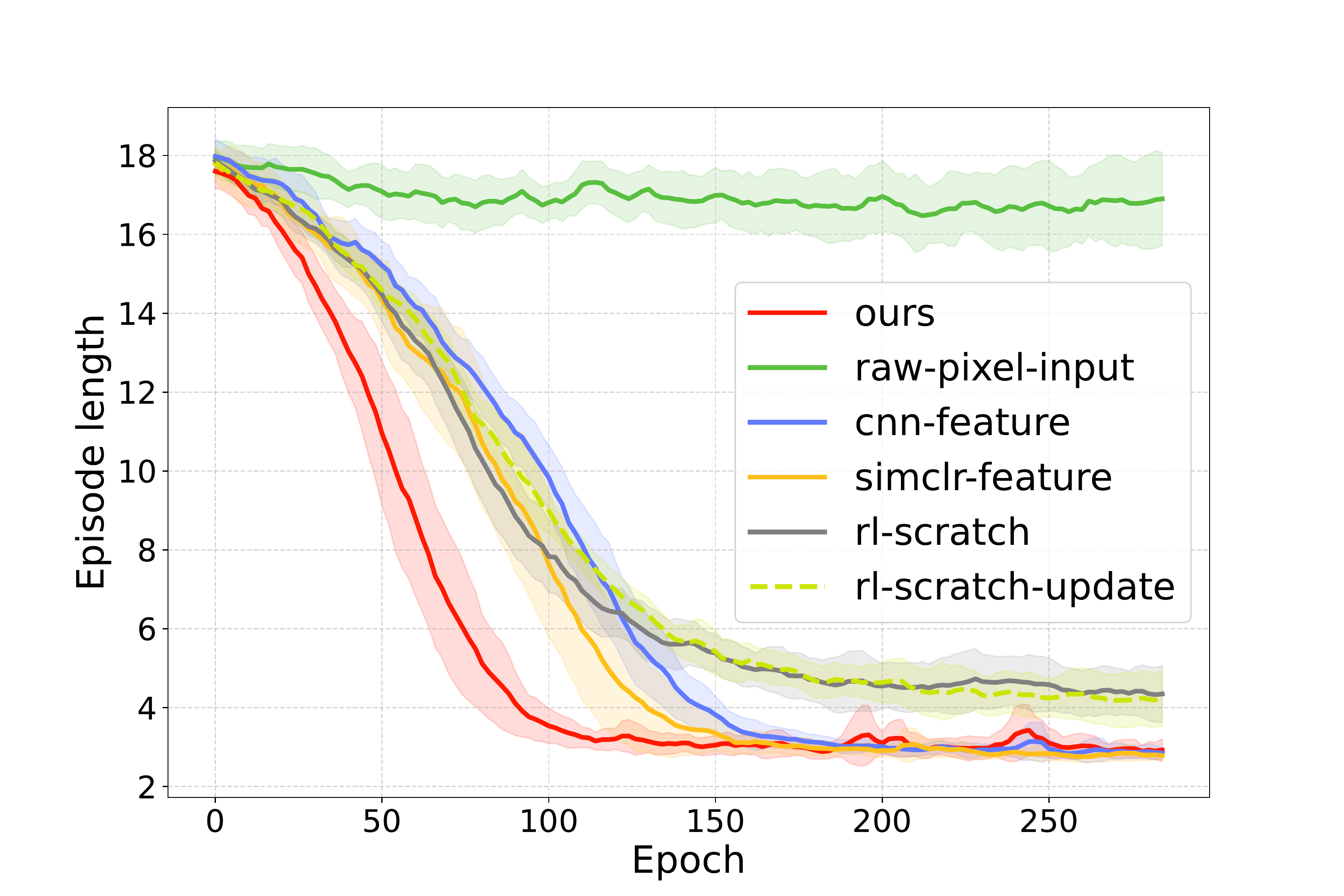}
    \caption{episode length of accomplishing the task}
    \label{fig:app-ep_len}
  \end{subfigure}
  \caption{Performance of new Listener trained with different inputs of the target state in the Object Placement task, with \textit{rl-scratch-update} added.}
  \label{fig:app-obj}
\end{figure}

In the Object Placement task, we also try to finetune the learned Speaker from the referential game. We do this by running the \textit{rl-scratch} baseline with Speaker initialized with the learned parameters. We call this setting \textit{rl-scratch-update}. The results are shown in Figure~\ref{fig:app-obj}. The performance of \textit{rl-scratch-update} is close to \textit{rl-scratch}, probably because the learned language is destroyed during the exploration of the new task training.

\end{document}